\documentclass[10pt,twocolumn,letterpaper]{article}

\usepackage{wacv}
\usepackage{times}
\usepackage{epsfig}
\usepackage{graphicx,subcaption}
\usepackage{amsmath}
\usepackage{amssymb}

\wacvfinalcopy 
\ifwacvfinal
\def\assignedStartPage{1} 
\fi

\ifwacvfinal
\usepackage[breaklinks=true,bookmarks=false]{hyperref}
\else
\usepackage[pagebackref=true,breaklinks=true,colorlinks,bookmarks=false]{hyperref}
\fi

\ifwacvfinal
\else
\fi

\begin{document}
	
	\title{R-MNet: A Perceptual  Adversarial Network for Image Inpainting}
	
\author{\parbox{16cm}{\centering
		{\large Jireh Jam$^1$, Connah Kendrick$^1$,Vincent Drouard$^2$, Kevin Walker$^2$, Gee-Sern Hsu$^3$ and \\Moi Hoon Yap$^1$}\\
		{\normalsize
			$^1$ Manchester Metropolitan University, Manchester, UK\\
			$^2$ Image Metrics Ltd, Manchester, UK\\
			$^3$ National Taiwan University of Science and Technology, Taipei, Taiwan}}
}
	
\maketitle

\begin{abstract}
	Facial image inpainting is a problem that is widely studied, and in recent years the introduction of Generative Adversarial Networks, has led to improvements in the field. Unfortunately some issues persists, in particular when blending the missing pixels with the visible ones. We address the problem by proposing a Wasserstein GAN combined with a new reverse mask operator, namely Reverse Masking Network (R-MNet), a perceptual  adversarial network for image inpainting. The reverse mask operator transfers the reverse masked image to the end of the encoder-decoder network leaving only valid pixels to be inpainted. Additionally, we propose a new loss function computed in feature space to target only valid pixels combined with adversarial training. These then capture data distributions and generate images similar to those in the training data with achieved realism (realistic and coherent) on the output images. We evaluate our method on publicly available dataset, and compare with state-of-the-art methods.  We show that our method is able to generalize to high-resolution inpainting task, and further show more realistic outputs that are plausible to the human visual system when compared with the state-of-the-art methods. https://github.com/Jireh-Jam/R-MNet-Inpainting-keras
\end{abstract}

\section{Introduction}
\label{sec:introduction}
Image restoration is achieved through the process of image inpainting, a technique initially performed by hand to restore images damaged by defects (e.g. cracks, dust, spots, scratches) to maintain image quality. Recently, image inpainting has taken a digital format, and is defined in computer vision as applying sophisticated algorithms that interpolate pixels for disocclusion, object removal and restoration of damaged images. Research in this area of study has been propelled by the increased demand for photo editing in mobile applications  ({\it e.g.} Snapseed, lightroom, pixlr express and flickr),  where modifications are done by erasing unwanted scenes and/or recovering occluded areas. Images contain visible structural and textural information which when distorted can be easily recognized by the human visual system. Maintaining image realism is therefore of utmost importance. 
Several methods have attempted to maintain image realism. These methods can be classified into two groups: Conventional and deep learning methods. Conventional methods approach image inpainting using texture synthesis techniques via mathematical equations, to obtain image statistics from surrounding pixel similarity for best fitting pixels to fill in missing regions caused by defects. 
These methods \cite{efros1999texture,bertalmio2000image,barnes2009patchmatch,criminisi2004region} use extended textures from local surrounding of similar pixels to fill in missing regions. Using patches of similar textures by Barnes et al. \cite{barnes2009patchmatch}, this technique can synthesize content. However, it lacks high-level semantic details and often generates structures that are non-realistic with repetitive patterns. 

In contrast, the second group of approaches (deep learning methods) \cite{pathak2016context,iizuka2017globally,yeh2017semantic,yang2017high,li2017generative,liu2018image,yan2018shift,yu2018generative,jam2020symmetric} uses generative neural networks to hallucinate missing content of an image based on encoding the semantic context of the image into feature space for realistic output by a decoder. 
This is done through convolutions which is an operation that extracts feature maps by evaluating the dot product between a kernel and each location of the input image.
The convolutional features are usually propagated channel-wise or through a fully connected layer to a decoder for reconstruction and may sometimes yield images: 1) that are overly smooth (blurry) and 2) with texture artefacts that lack edge preservation. Usually the lack of meaningful semantic content on inpainted images by the state of the art can result from applying the loss function on the entire image, which evaluates the error of the image as a whole, instead of focusing on predicted pixels in masked regions.
In this paper, we propose a novel generative neural network, namely R-MNet that predicts missing parts of an image and preserves its realism with complete structural and textural information. Our algorithm takes into account the global semantic structure of the image and predicts fine texture details for the missing parts that are consistent and realistic to the human visual system. 

To enable and enhance the reversed masking mechanism, we propose a reverse-mask loss in feature space that measures the distance between predicted output of the mask regions and the corresponding original pixel areas of the mask on the input image. To achieve this, we use features from VGG-19 pre-trained on ImageNet by Simonyan et al. \cite{simonyan2015very}. We revisit Context-Encoder by Pathak et al. \cite{pathak2016context} and without bias, we make significant changes that overcome the limitations encountered by existing state of the art. We design an architecture that focuses on the missing pixel values, to extract features and encode them in latent space in an end-to-end fashion. 
In summary, our main contributions are:
\begin{itemize}
	\item We propose an end-to-end Reverse Masking Wasserstein GAN image inpainting framework (R-MNet) with improved performance when compared to the state of the art.
	\item The proposed reverse masking technique can improve the quality of inpainting results by applying the reversed mask on the masked-image as target regions for inpainting.
	\item  A perceptually motivated new combination loss function defined using high level features to target missing pixels to train the novel R-MNet to produce images of high visual quality. 
\end{itemize}

Our approach has achieved high-quality results. The output images when compared with the state of the art have more coherent texture and structures similar to the original images without any post processing.
\section{Related Work}
\label{sec:related}
Recent work on image inpainting is predominantly focused on deep learning methods due to their ability to capture distribution of high-dimensional data (images). The use of neural network on image inpainting was first approached by Jain et al. \cite{jain2007supervised} as an image denoising task formulated with parameter learning for backpropagation with the noisy image as the learning problem. However, restricted to one colour channel with a single input layer and limited to noise, this model was extended to inpainted images but required substantial computation. 

Pathak et al. \cite{pathak2016context} introduced the use of adversarial training to image inpainting and propose the context-encoder that combines pixel-wise reconstruction loss with adversarial loss by Goodfellow et al. \cite{goodfellow2014generative} to predict missing pixel values on images. Iizuka et al. \cite{iizuka2017globally}, improved upon the Pathak et al. \cite{pathak2016context} and proposed the use of two discriminators; a local discriminator to assess the consistency of predicted pixels and a global discriminator to assess the coherency of the entire image. Yang et al. \cite{yang2017high} introduced a combined optimisation framework, a multi-scale neural synthesis approach with a constraint to preserve global and local texture during pixel-wise prediction of missing regions. Yeh et al. \cite{yeh2017semantic} proposed to search closest encoding with context and prior loss combined with adversarial training for the reconstruction of images in latent space. Liu et al. \cite{liu2018image} propose partial convolutions with automatic mask updating combined with re-normalised convolutions to target only valid (missing) pixel prediction. Yan et al. \cite{yan2018shift} added a special shift connection layer that serves as a guidance loss to the U-NET \cite{ronneberger2015u} architecture for deep feature rearrangement of sharp structures and fine texture details.

Yu et al. \cite{yu2018generative} introduced a coarse-to-fine network with contextual attention layer that replaces the Poisson blending post-processing step in \cite{iizuka2017globally}.
Wang et al. \cite{wang2019laplacian} proposed a Laplacian pyramid approach, supported by residual learning \cite{he2016deep} that propagates high frequency details to predict missing pixels at different resolutions. Huang et al. \cite{huang2019image} proposed an image completion network-based adversarial loss combined with $L_{1}$ and Structural Similarity Index Measure (SSIM) \cite{wang2004image} to improve on the structural texture and authenticate the reconstructed image.
Zeng et al. \cite{zeng2019learning} proposed cross-layer attention and pyramid filling mechanisms to learn high level semantic features from region affinity to fill pixels of missing regions in a pyramid fashion. Li et al. \cite{li2019progressive} revisited partial convolution \cite{liu2018image} and introduced a visual structure reconstruction layer to incorporate the structural information in the reconstructed feature map.
Liu et al. \cite{liu2019coherent} proposed a coherent semantic attention layer on a dual-network embedded in the encoder as a refinement mechanism. Ren et al. \cite{ren2019structureflow} introduced a two-stage model for structure reconstruction and texture generation with the help of appearance flow in the texture generator to yield image details.

In 2019, Yu et al. \cite{yu2019free} proposed the use of gated convolutions in image inpainting that automatically learns soft mask from data. This network \cite{yu2019free} combines gated convolutions with an attention layer to automatically learn a dynamic feature mechanism for each channel at each spatial location. Guo et al. \cite{guo2019progressive} proposed the use of partial convolutions in residual network with feature integration determined by dilation step that assigns several residual blocks as a one dilation strategy combined with a step loss for intermediate restoration.
Wang et al. \cite{wang2019musical} used a multi-scale attention module that utilizes background content. Xie et al. \cite{xie2019image} introduced bidirectional attention maps that target irregular hole filling within the decoder. Wang et al \cite{wang2020multistage} reused partial convolutions combined with multi-stage attention that computes background and foreground scores from pixel-wise patches. within the decoder to improve quality during the reconstruction of irregular masked regions.
However, these methods have a common practice of using multiple networks as part of the generator or discriminator and expensive post-processing to perform an image inpainting task and do not consider using visible information that targets the missing pixels.

\section{Proposed Framework}
\label{sec:framework}
In this section, we present our R-MNet for solving image inpainting tasks. We employ as base a Wasserstein Generative Adversarial Network (WGAN), using the encoding-decoding architecture of WGAN networks we introduce our new reverse masking operator that enforces the model to target missing pixels thus allowing the network to recover hidden part of an image while keeping the visible one. We also defined a new loss function namely reverse masking loss build around this reverse masking operator.

\subsection{Network Architecture}
\label{subsec:architecture}
As mentioned previously, R-MNet is build using a GAN as base architecture. GANs have been previously used in image inpainting as they are able to generate missing pixels, unfortunately this often leads to the introduction of blurriness and/or artefact effects. Recent works by Liu et al. \cite{liu2018image}, Guo et al. \cite{guo2019progressive} and Yu et al \cite{yu2019free} try to solve this problem by using partial convolution and gated convolutions. While these two approaches aim to target more efficiently missing pixels we found that they do not fully reduced the aberrations.
Our aim through the reverse masking operator is to better target missing region in the image while keeping visible pixels intact. 

First, we need to define some generic terminology that will be used through the rest of the paper. We define the source image as $\mathcal{I}$, the mask as $M$ and the reversed masked $M_r = 1 - M$. The masked input image $\mathcal{I}_M$ is obtained as follows:
\begin{equation}
\mathcal{I}_M = \mathcal{I} \odot M
\end{equation}
where $\odot$ is the element wise multiplication operator. 

\begin{figure*}
	\centering
	\includegraphics[width=0.85\linewidth]{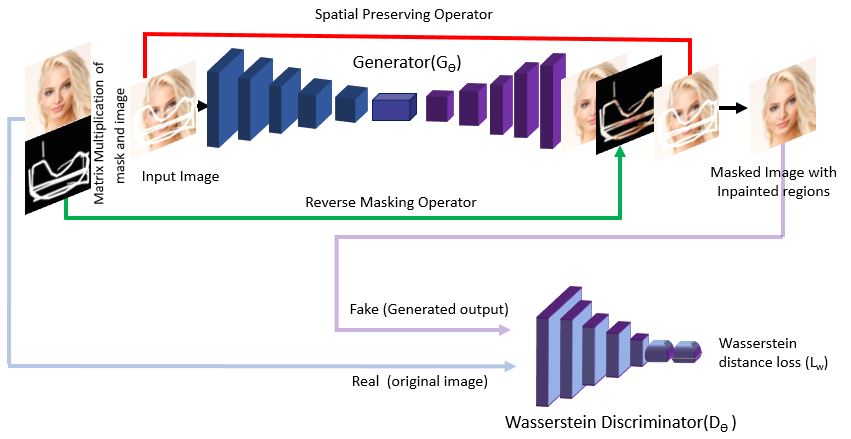} 
	\caption{
		An overview of R-MNet architecture at training showing the spatial preserving operation  and reverse-masking mechanism.}
	\label{fig:reverse-framework}
\end{figure*}

Our network architecture is designed to have a generator ($G_\theta$) and a Wasserstein Discriminator ($D_\theta$). Our generator is designed with convolutional and deconvolutional (learnable up-sampling ) layers. The convolutional layers encode features in latent space during convolution. These layers are blocks of convolution with filter size of 64 and the kernel size set to $5\times5$ with a dilation rate of 2 and Leaky-ReLU, $\alpha$=0.2. We included dilated convolution to widen the receptive field to capture fine details and textural information. The convolutional feature maps obtained in each layer are the input to the next layer after rectification and pooling. We use Maxpooling to reduce variance and computational complexity by extracting important features like edges, and keep only the most present features. We include in our learnable up-sampling layers, reflection padding on a kernel size that is divisible by the stride (K-size=4, stride=2), and bilinear interpolation to resize the image, setting the up-sampling to a high-resolution, and through a tanh function output layer. The goal of setting up the decoder in this way is to ensure that any checker-board artefacts \cite{rosebrock2019deep} associated with the inpainted regions on the output image are cleaned and consistent with details outside the region. This technique is equivalent to sub-pixel convolution achieved in \cite{shi2016real}. We include specifically the WGAN adopted from \cite{arjovsky2017wasserstein} that uses the Earth-Mover distance, as part of our network to compare generated and real distributions of high-dimensional data. 
The generator will produce a predicted image $\mathcal{I}_pred = G_{\theta}(\mathcal{I}_M)$. Using our reversed masked operator we obtain $M_r$ and combined it with $\mathcal{\mathcal{I}}_{pred}$ to produce predicted masked area image:
\begin{equation}
\mathcal{I}_{Mpred} = \mathcal{I}_{pred} \odot M_r
\end{equation}
The overall architecture is shown in Figure~\ref{fig:reverse-framework}. 
By using this approach, our model predicts only regions with missing pixels which are consistent with surrounding pixels close to border regions of the original image. This results in high-quality reconstructed images that match the natural texture and structure of the original images that are visually plausible with preserved realism. 

\subsection{Loss Function} 
\subsubsection{Generator Loss Function}
We define our generator loss function $\mathcal{L}_{G}$ to evaluate two aspect of the predicted image: the quality of missing pixels area and, the whole image perceptual quality. Building $\mathcal{L}_{G}$ around these two metrics will ensure that the generator produces accurate missing pixels that they will blend nicely with the visible pixels. State-of-the-art methods \cite{gatys2016neural,johnson2016perceptual,liu2018image,liu2019coherent,guo2019progressive} contribute to style transfer and image inpainting have used feature space instead of pixel space to optimize network. Using feature space encourages adversarial training to generate images with similar features, thus achieving more realistic results. Our new combination of loss function is computed based on feature space. We achieve this by utilizing pre-trained weights from the VGG-19 model trained on ImageNet \cite{krizhevsky2012imagenet}. We extract features from block3-convolution3 and compute our loss function using Mean Square Error (MSE) \cite{pathak2016context} as our base. Instead of using pixel-wise representations, we use extracted features and compute the squared difference applied to the input and output of our loss model as our perceptual loss ($\mathcal{L}_{p}$), which is similar to \cite{johnson2016perceptual}, as in equation \ref{eq:ploss-function}:
\begin{equation}
\mathcal{L}_{p} = \frac{1}{\kappa} \sum_{i \in \phi}(\phi_i[\mathcal{I}] - \phi_i[\mathcal{I}_{pred}])^2
\label{eq:ploss-function}
\end{equation}
where $\kappa$ is the size of $\phi$ (output from block3-convolution3 of VGG19), $\phi_i[I]$ is the feature obtained by running the forward pass of VGG19 using $I$ as input and $\phi_i[\mathcal{I}_{pred}]$ is the feature obtained by running the forward pass on the output of the generator $G_{\theta}[I_M]$.

We define our reversed mask-loss ($\mathcal{L}_{rm}$) on the same bases as MSE, but targeting only valid features created by the mask region for reconstruction. Our reversed mask loss compares the squared difference for corresponding pixels specific for regions created by the mask on the image and the reconstructed pixels of the masked-image. We use the reversed mask ($M_{r}$) and the original image ($X_{input}$) to obtain $\mathcal{L}_{rm}$, where
\begin{equation}
\mathcal{L}_{rm} = \frac{1}{\kappa} \sum_{i \in \phi} (\phi_{i}[\mathcal{I}_{Mpred}] -\phi_{i}[\mathcal{I} \odot M_r])^2 
\label{eq:mloss-function}
\end{equation}
Finally by linearly combining $\mathcal{L}_{p}$ and $\mathcal{L}_{rm}$ we obtain the generator loss function:
\begin{equation}
\mathcal{L}_{G} = (1-\lambda)\mathcal{L}_{p} + \lambda\mathcal{L}_{rm},
\label{eq:valid_loss}
\end{equation}
where $\lambda\in [0 \; 1] $, to allow an optimal evaluation of features by minimising the error on the missing region to match predictions comparable to the ground-truth.

\subsubsection{Discriminator Loss Function}
Since we train our network with Wasserstein distance loss function ($\mathcal{L}_{w}$), we define this in equation~\ref{eq:W-distance-loss-function}.
\begin{equation}
\mathcal{L}_{w} = E_{\mathcal{I}\sim P_{x}}[D_{\theta}(\mathcal{I})]- E_{\mathcal{I}_{pred}\sim P_{z}}[D_{\theta}(\mathcal{I}_{pred})]  
\label{eq:W-distance-loss-function}
\end{equation}
Here the first term is the probability of real data distribution and the second term is the generated data distribution. 

\subsection{Reverse Mask}
\begin{figure*}
	\centering
	\includegraphics[width=.85\linewidth]{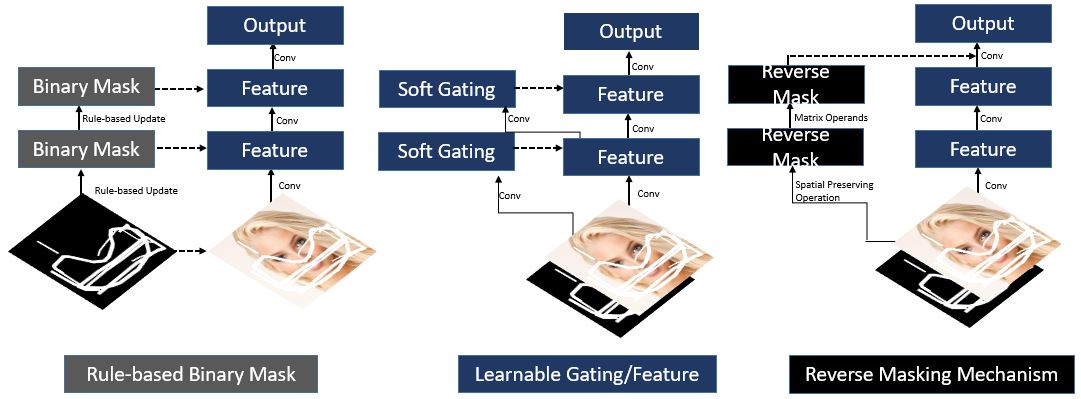} 
	\caption{
		Illustration of partial convolution (left) and gated convolution (middle) and Reverse-masking (right).}
	\label{fig:illustration}
\end{figure*}

We discuss the advantages of our approach using reverse mask operator compared to Partial Convolution (PConv) and Gated Convolution (GC), two approaches previously used for image inpainting. All three methods are summarized in Figure~\ref{fig:illustration}. The process in partial convolution layers takes in both the image and mask to produce features with a slightly filled mask. Each partial convolutional layer has a mask which if renormalised focuses on valid pixels and an automatic mask update for the next layer. With more partial convolution layers, the mask region gets smaller and smaller, which can disappear in deeper layers and revert all mask values to ones. With gated convolutions , the convolutions automatically learn soft mask from data, extracting features according to mask regions. Each convolution block learns dynamic features for each channel at each spatial location and pass it through different filters. In this process, the output features goes through an activation mechanism (ReLU) while gating values (between zeros and ones) goes through sigmoid function. The gating values show learnable features with semantic segmentation and highlighted mask regions as a sketch in separate channels to generate inpainting results. This network requires a substantial amount of CPU/GPU memory to run the gating scheme. Our proposed reverse mask forces the convolutions to subtract the non-corrupt areas through the reverse masking mechanism ensuring final predictions of the missing regions, with the help of the reverse mask loss, forcing the network via the backward pass to focus on the predictions of the missing regions yielding more plausible outcomes.

\section{Experiment}
\label{sec:experiment}
In this section, we introduce datasets used in this work and present our implementation. We evaluate our approach qualitatively and quantitatively, demonstrating how effective our method is in performing image inpainting without any post-processing mechanism. 
\begin{figure}   
	\centering
	\begin{subfigure}[b]{0.15\linewidth}        
		\centering
		\includegraphics[width=\linewidth]{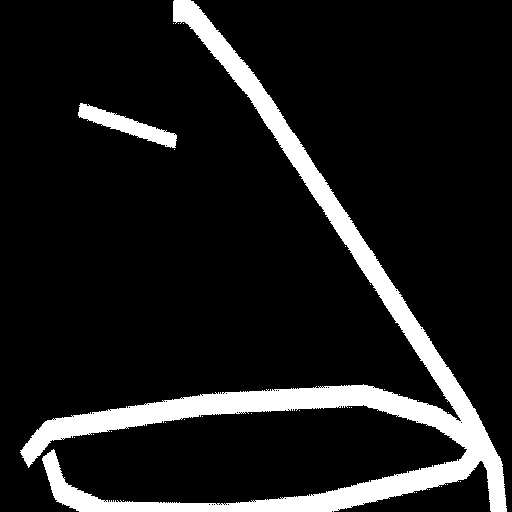}
	\end{subfigure}
	\begin{subfigure}[b]{0.15\linewidth}        
		\centering
		\includegraphics[width=\linewidth]{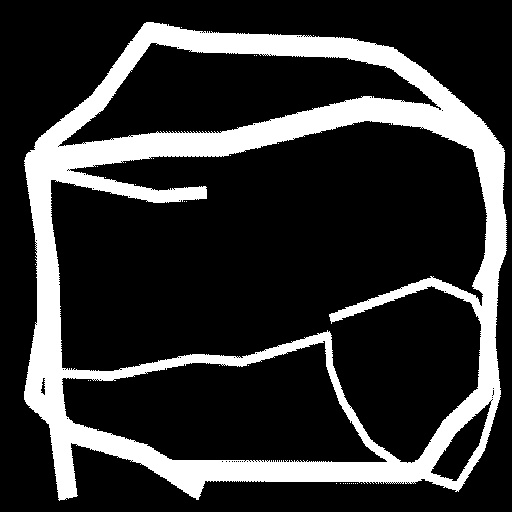}
	\end{subfigure}
	\begin{subfigure}[b]{0.15\linewidth}        
		\centering
		\includegraphics[width=\linewidth]{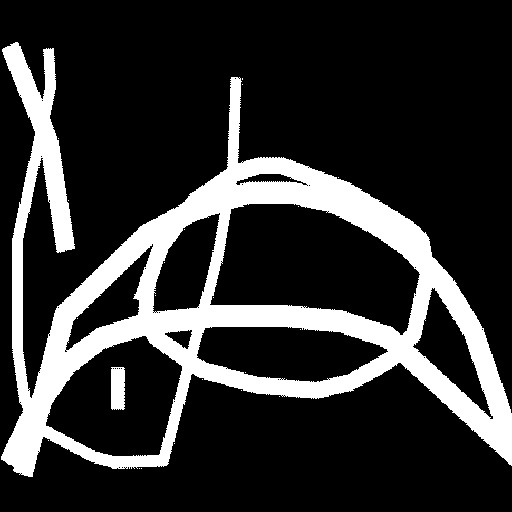}
	\end{subfigure}
	\begin{subfigure}[b]{0.15\linewidth}        
		\centering
		\includegraphics[width=\linewidth]{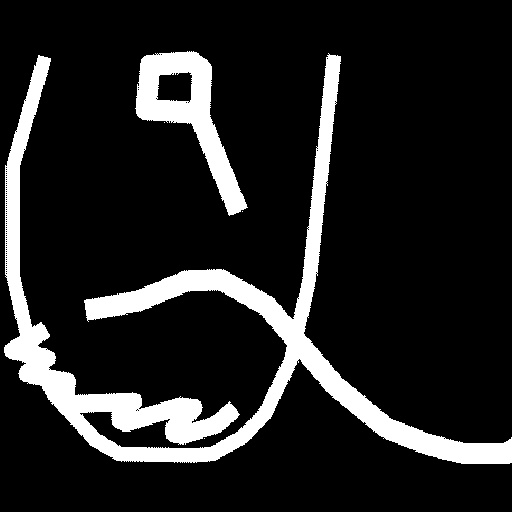}
	\end{subfigure}
	\begin{subfigure}[b]{0.15\linewidth}        
		\centering
		\includegraphics[width=\linewidth]{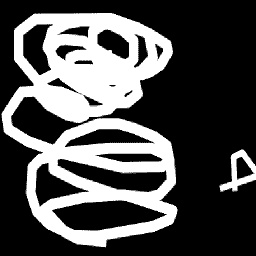}
	\end{subfigure}
	\begin{subfigure}[b]{0.15\linewidth}        
		\centering
		\includegraphics[width=\linewidth]{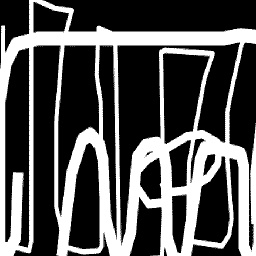}
	\end{subfigure}
	\caption[Optional caption for list of figures 5-8]{Sample images from Quick-Draw Dataset by Iskakov et al. \cite{iskakov2018semi}}
	\label{fig:Masks}
\end{figure}
\subsection{Datasets}
We use three publicly available datasets to evaluate our method: CelebA-HQ \cite{karras2018progressive}, Places2 \cite{zhou2018places} and the Paris Street View \cite{doersch2012makes}. For masking, we use the Quick Draw mask dataset \cite{iskakov2018semi} shown on Figure~\ref{fig:Masks}, which contains 50,000 train and 10,000 test sets designed based on 50 million human drawn strokes in vector format combined to form irregular shapes (patterns) \cite{iskakov2018semi}  of size $512 \times 512$ pixels.
During training and testing, we randomly select the masks from various sets and resize each to $256 \times 256$. We used the CelebA-HQ curated from the CelebA dataset \cite{liu2015deep} by Karras et al. \cite{karras2018progressive}. From the 30,000 high quality images of $1024 \times 1024$, $512 \times 512$ and $128 \times 128$ resolutions, we followed the state-of-the-art procedures \cite{liu2018image} and split our dataset into 27,000 training and 3,000 testing set.  The Places2 \cite{zhou2018places} dataset contains more than 10 million images with more than 400 unique scene categories with 5,000 to 30,000 train images. We split the training and testing set according to the state-of-the-art \cite{pathak2016context} and trained our network to understand scene category. 
The Paris Street View has 14,900 training images and 100 validation images. We use the same testing set as described in the state-of-the-art \cite{pathak2016context} for our experiment.

\subsection{Implementation}
We use the Keras library with TensorFlow backend to implement our model. Our choice of datasets matches the state of the art \cite{liu2018image,liu2019coherent,pathak2016context,liu2019coherent} with similar experimental settings. We resize all images and masks using OpenCV library interpolation function $ \verb|INTER_AREA|$ to $256 \times 256 \times 3$ and  $256 \times 256 \times 1$ respectively. 
We use the Adam optimizer\cite{kingma2015adam} with learning rate of $10^{-4}$, $\beta=0.9$ for $G_{\theta}$ and $10^{-12}$, $\beta=0.9$ for $D_{\theta}$. We train our model with a batch size of 5 on NVIDIA Quadro P6000 GPU machine, on Places2 and Paris Street View. We use NVIDIA GeForce GTX 1080 Ti Dual GPU machine on CelebA-HQ  dataset high-resolution images. It takes 0.193 seconds to predict missing pixels of any size created by binary mask on an image, and 7 days to train 100 epochs of 27,000 high-resolution images.

\section{Results}
\label{sec:results}
To evaluate the performance of the proposed method, we compare R-MNet with three other methods on the same settings for image size, irregular holes and datasets. Our experiments include 
\begin{itemize}
	\item \textbf{CE}: Context-Encoder by Pathak et al. \cite{pathak2016context}
	\item \textbf{PConv}: Partial Convolutions by Liu et al. \cite{liu2018image}
	\item \textbf{GC}: Free-Form image inpainting with Gated Convolutions by Yu et al. \cite{yu2019free}	
	\item \textbf{R-MNet-0.1}: R-MNet using $\boldsymbol\ell_{rm}$ when {$\lambda$= \textbf{0.1}}, our proposed method.
	\item \textbf{R-MNet-0.4}: R-MNet using $\boldsymbol\ell_{rm}$ when  {$\lambda$= \textbf{0.4}}, our proposed method.
\end{itemize}
\begin{figure*}
	\centering
	\begin{subfigure}[b]{0.14\linewidth}        
		\centering
		\includegraphics[width=\linewidth]{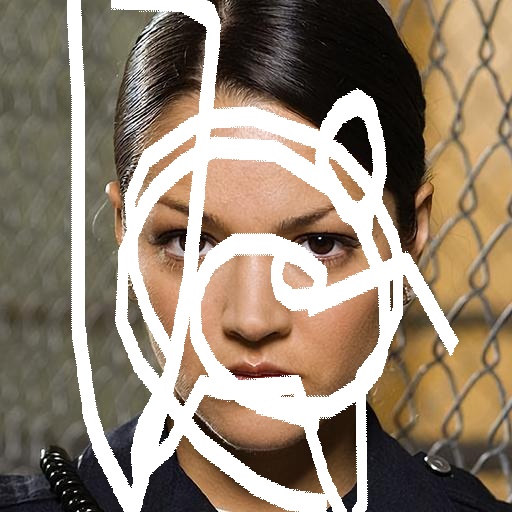}
	\end{subfigure}
	\begin{subfigure}[b]{0.14\linewidth}        
		\centering
		\includegraphics[width=\linewidth]{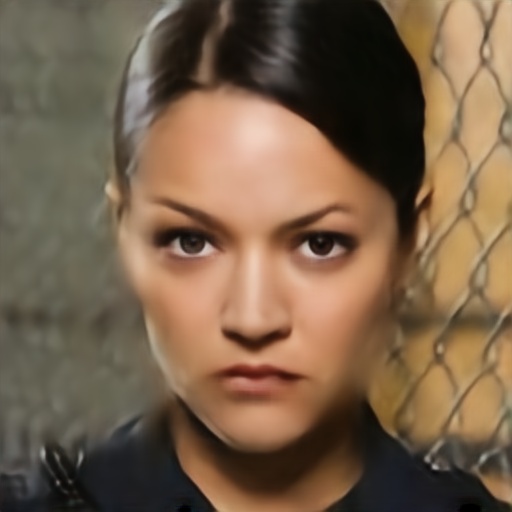}
	\end{subfigure}
	\begin{subfigure}[b]{0.14\linewidth}        
		\centering
		\includegraphics[width=\linewidth]{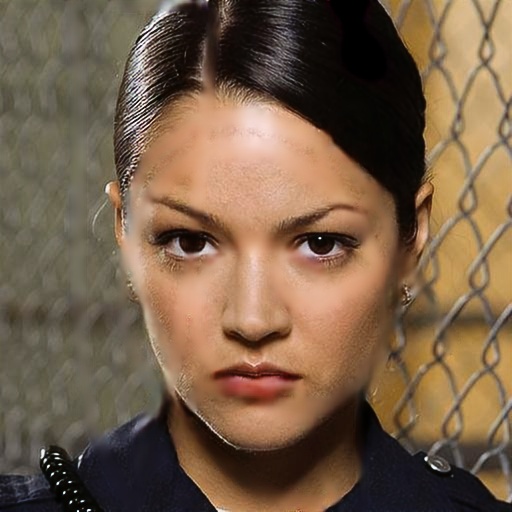}
	\end{subfigure}
	\begin{subfigure}[b]{0.14\linewidth}        
		\centering
		\includegraphics[width=\linewidth]{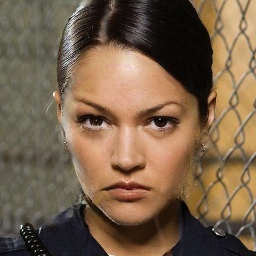}
	\end{subfigure}
	\begin{subfigure}[b]{0.14\linewidth}        
		\centering
		\includegraphics[width=\linewidth]{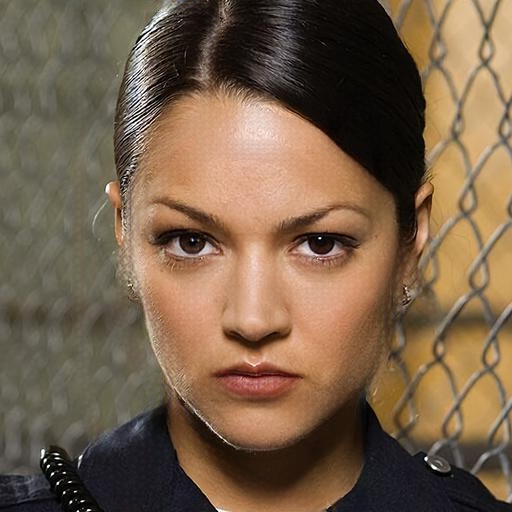}
	\end{subfigure}
	\begin{subfigure}[b]{0.14\linewidth}        
		\centering
		\includegraphics[width=\linewidth]{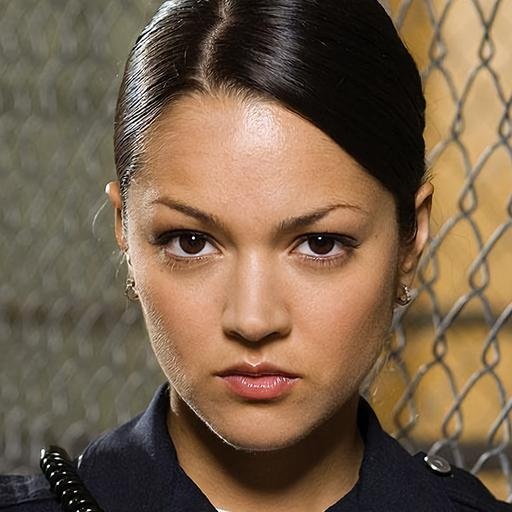}
	\end{subfigure}
	\\
	\begin{subfigure}[b]{0.14\linewidth}        
		\centering
		\includegraphics[width=\linewidth]{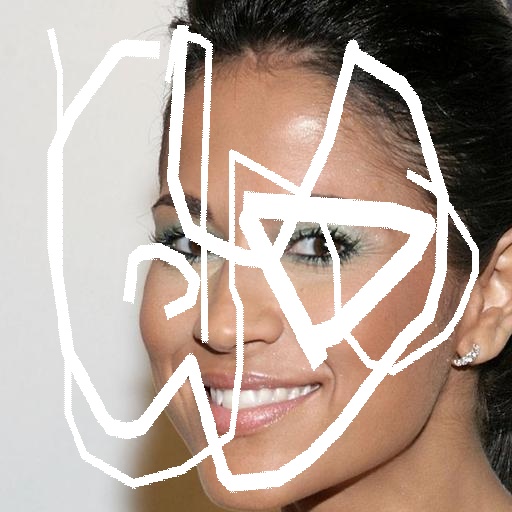}
	\end{subfigure}
	\begin{subfigure}[b]{0.14\linewidth}        
		\centering
		\includegraphics[width=\linewidth]{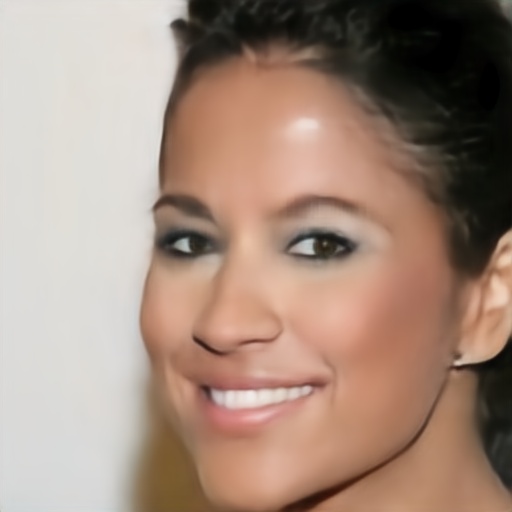}
	\end{subfigure}
	\begin{subfigure}[b]{0.14\linewidth}        
		\centering
		\includegraphics[width=\linewidth]{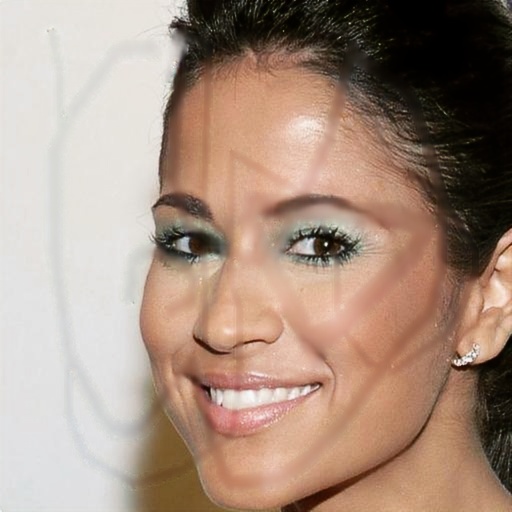}
	\end{subfigure}
	\begin{subfigure}[b]{0.14\linewidth}        
		\centering
		\includegraphics[width=\linewidth]{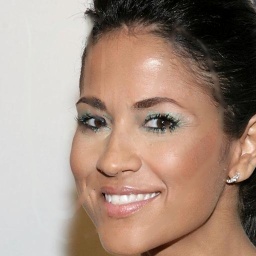}
	\end{subfigure}
	\begin{subfigure}[b]{0.14\linewidth}        
		\centering
		\includegraphics[width=\linewidth]{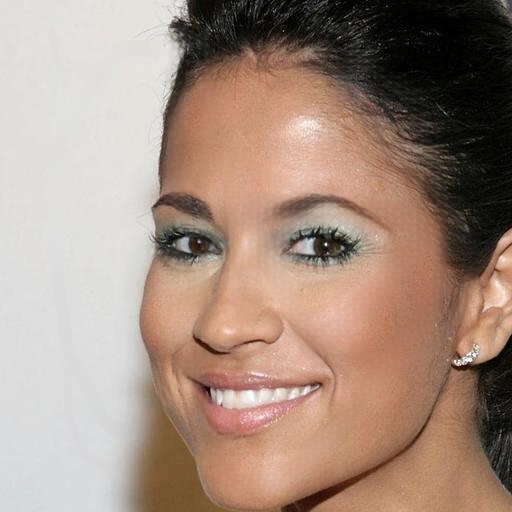}
	\end{subfigure}
	\begin{subfigure}[b]{0.14\linewidth}        
		\centering
		\includegraphics[width=\linewidth]{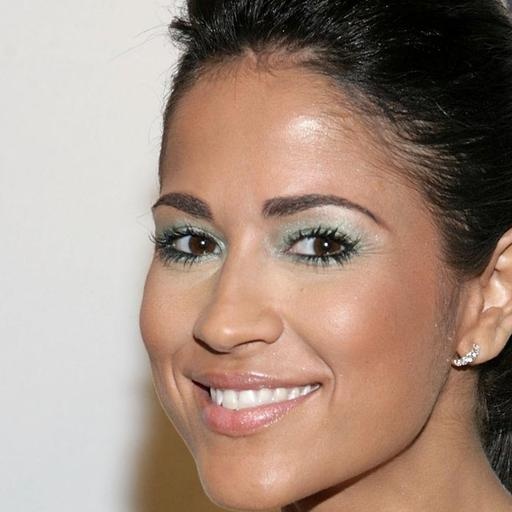}
	\end{subfigure}
	\\	
	\begin{subfigure}[b]{0.14\linewidth}        
		\centering
		\includegraphics[width=\linewidth]{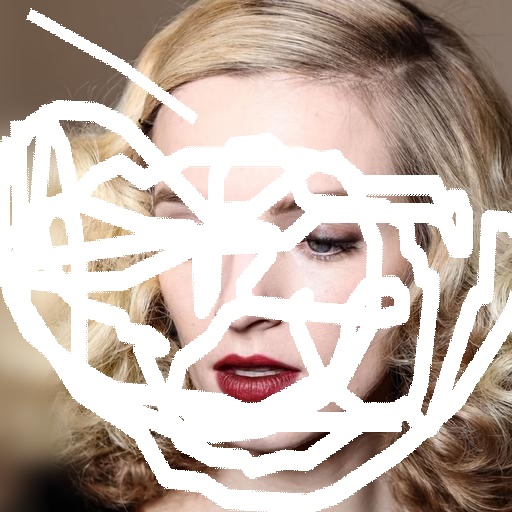}
		\caption{Masked}
	\end{subfigure}
	\begin{subfigure}[b]{0.14\linewidth}        
		\centering
		\includegraphics[width=\linewidth]{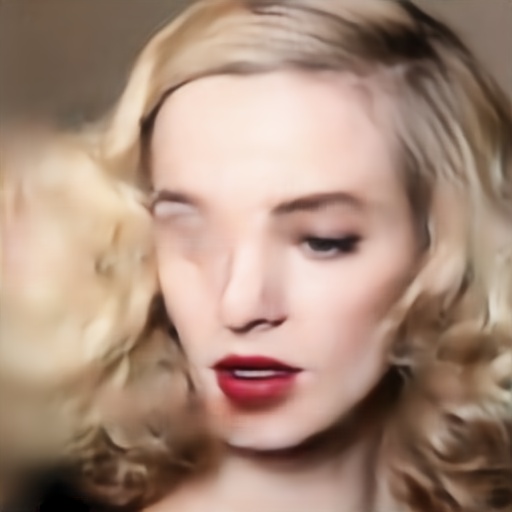}
		\caption{CE \cite{pathak2016context}}
	\end{subfigure}
	\begin{subfigure}[b]{0.14\linewidth}        
		\centering
		\includegraphics[width=\linewidth]{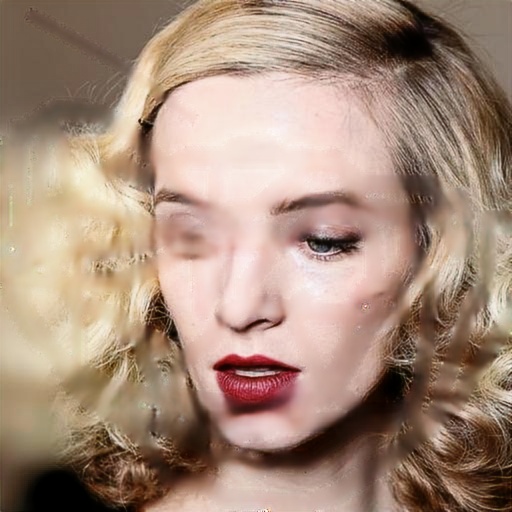}
		\caption{PConv \cite{liu2018image}}
	\end{subfigure}
	\begin{subfigure}[b]{0.14\linewidth}        
		\centering
		\includegraphics[width=\linewidth]{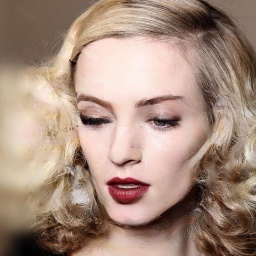}
		\caption{GC \cite{yu2019free}}
	\end{subfigure}
	\begin{subfigure}[b]{0.14\linewidth}        
		\centering
		\includegraphics[width=\linewidth]{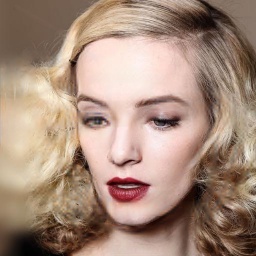}
		\caption{R-MNet}
	\end{subfigure}
	\begin{subfigure}[b]{0.14\linewidth}        
		\centering
		\includegraphics[width=\linewidth]{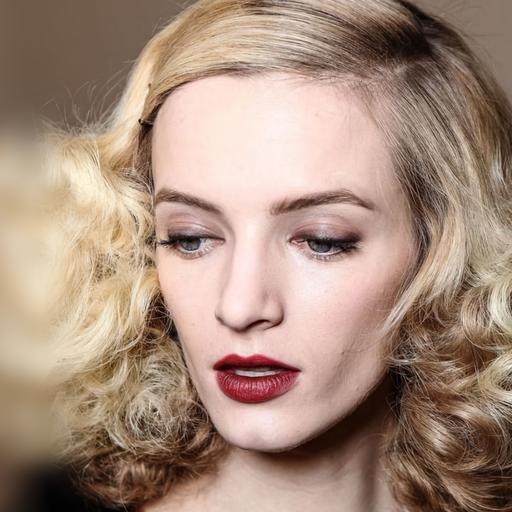}
		\caption{GT}
	\end{subfigure}
	\caption{Visual comparison of the inpainted results by  \textbf{CE}, \textbf{PConv}, \textbf{GC} and \textbf{R-MNet} on CelebA-HQ \cite{liu2018image} where Quick Draw dataset \cite{iskakov2018semi} is used as masking method using mask hole-to-image ratios [0.01,0.6].}
	\label{fig:Comparedresults}
\end{figure*}

\subsection{Qualitative Comparison}
We carried out experiments based on similar implementations by Liu et al. \cite{liu2018image}, Pathak et al. \cite{pathak2016context} and pre-trained model for the state of the art \cite{yu2019free} and compared our results.
\begin{figure}   
	\centering
	\begin{subfigure}[b]{0.26\columnwidth}        
		\centering
		\includegraphics[width=\linewidth]{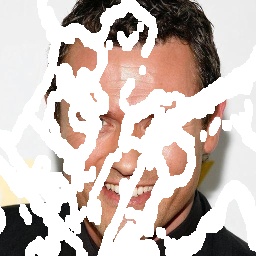}
	\end{subfigure}
	\begin{subfigure}[b]{0.26\columnwidth}        
		\centering
		\includegraphics[width=\linewidth]{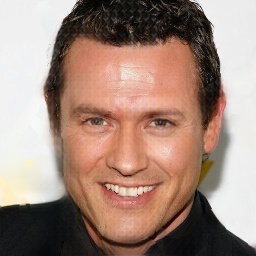}
	\end{subfigure}
	\begin{subfigure}[b]{0.26\columnwidth}        
		\centering
		\includegraphics[width=\linewidth]{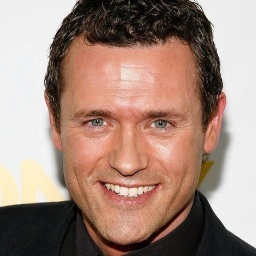}
	\end{subfigure}\\
	\begin{subfigure}[b]{0.26\linewidth}        
		\centering
		\includegraphics[width=\linewidth]{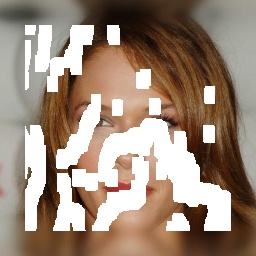}
	\end{subfigure}
	\begin{subfigure}[b]{0.26\linewidth}        
		\centering
		\includegraphics[width=\linewidth]{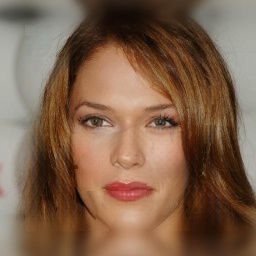}
	\end{subfigure}
	\begin{subfigure}[b]{0.26\linewidth}        
		\centering
		\includegraphics[width=\linewidth]{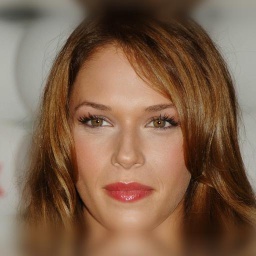}
	\end{subfigure}\\ 
	\begin{subfigure}[b]{0.26\linewidth}        
		\centering
		\includegraphics[width=\linewidth]{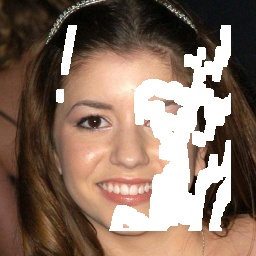}
		\caption{Masked}
	\end{subfigure}
	\begin{subfigure}[b]{0.26\linewidth}        
		\centering
		\includegraphics[width=\linewidth]{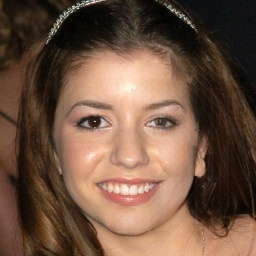}
		\caption{R-MNet}
	\end{subfigure}
	\begin{subfigure}[b]{0.26\linewidth}        
		\centering
		\includegraphics[width=\linewidth]{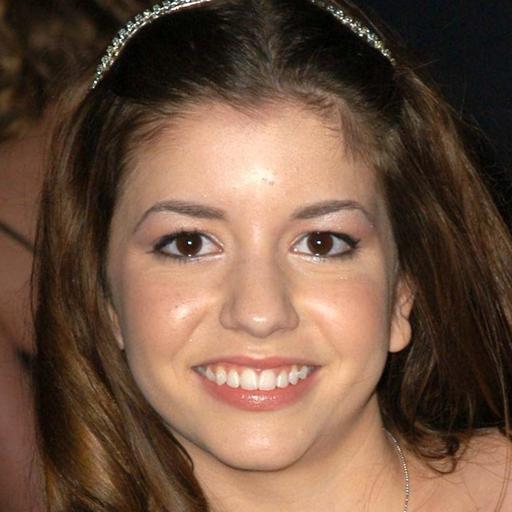}
		\caption{GT}
	\end{subfigure}   
	\caption{Results of image inpainting using R-MNet-0.4 on CelebA-HQ Dataset \cite{liu2018image} with Nvidia Mask dataset \cite{liu2018image} used as masks, where images in column (a) are the masked-image generated using the Nvidia Mask dataset \cite{liu2018image}; images in column (b) are the results of inpainting by our proposed method; and images in column (c) are the ground-truth.}
	\label{fig:results}
\end{figure}

\begin{figure}[ht]  
	\centering
	\begin{subfigure}[b]{0.26\linewidth}        
		\centering
		\includegraphics[width=\linewidth]{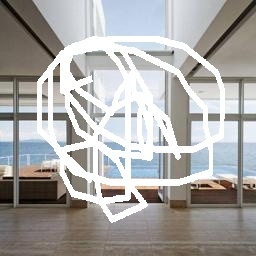}
	\end{subfigure}
	\begin{subfigure}[b]{0.26\linewidth}        
		\centering
		\includegraphics[width=\linewidth]{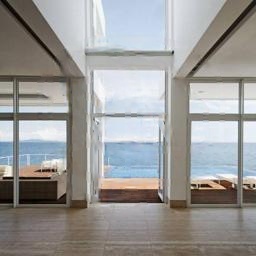}
	\end{subfigure}
	\begin{subfigure}[b]{0.26\linewidth}        
		\centering
		\includegraphics[width=\linewidth]{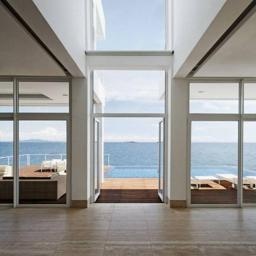}
	\end{subfigure}
	\\
	\begin{subfigure}[b]{0.26\linewidth}        
		\centering
		\includegraphics[width=\linewidth]{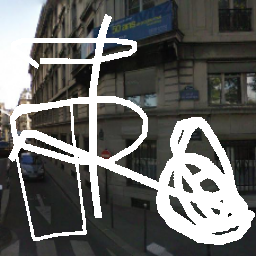}
	\end{subfigure}
	\begin{subfigure}[b]{0.26\linewidth}        
		\centering
		\includegraphics[width=\linewidth]{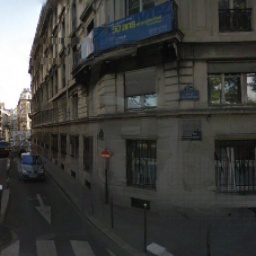}
	\end{subfigure}
	\begin{subfigure}[b]{0.26\linewidth}        
		\centering
		\includegraphics[width=\linewidth]{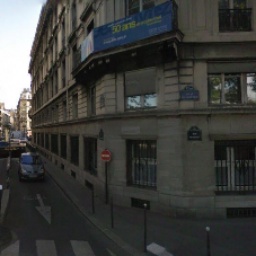}
	\end{subfigure}
	\\
	\begin{subfigure}[b]{0.26\linewidth}        
		\centering
		\includegraphics[width=\linewidth]{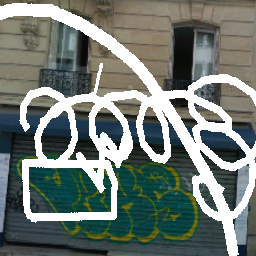}
		\caption{Masked}
	\end{subfigure}
	\begin{subfigure}[b]{0.26\linewidth}        
		\centering
		\includegraphics[width=\linewidth]{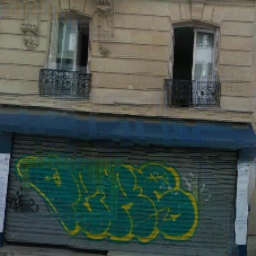}
		\caption{R-MNet}
	\end{subfigure}
	\begin{subfigure}[b]{0.26\linewidth}        
		\centering
		\includegraphics[width=\linewidth]{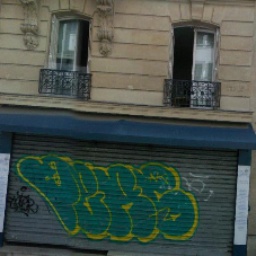}
		\caption{GT}
	\end{subfigure}
	\caption{Results of image inpainting using R-MNet-0.4 on Places2 \cite{zhou2018places} and Paris Street View \cite{doersch2012makes}, where images in column (a) are the masked-image generated using the Quick-Draw dataset \cite{iskakov2018semi}; images in column (b) are the results of inpainting by our proposed method; and images in column (c) are the ground-truth.}
	\label{fig:Paris-results}
\end{figure}

For Places2 dataset, we randomly select 10,000 training samples to match \cite{pathak2016context} and use the same number of test samples to evaluate our model. We show our results on Figure~\ref{fig:Comparedresults}.
For CelebA-HQ, we downloaded pre-trained models for the state of the art \cite{yu2019free} and compared our results.
Based on visual comparison, our model shows realistic and coherent output images. Observing from Figure~\ref{fig:Comparedresults}, other models fail to yield images with structural and textural content as the images are either blurry or fail due to the image-to-hole ratio increase with arbitrary mask. 

\subsection{Quantitative Comparison}
To statistically understand the inpainting performance, we quantify our model with the state of the art and compared our results, based on four classic metrics: Frechet Inception Distance (FID) by Heusel et al. \cite{heusel2017gans}, Mean Absolute Error (MAE), Peak Signal to Noise Ratio (PSNR), and SSIM \cite{wang2004image}. The FID measures the quality of reconstructed images by  calculating the distance between feature vectors of ground-truth image and reconstructed images. The other metrics (MAE, PSNR, SSIM) evaluate at pixel and perceptual levels respectively. The results in Table~\ref{table:result1} are evaluated based on masks with various test hole-to-image area ratios ranging from [0.01,0.6], on a test set of 3000 images from the CelebA-HQ.

\setlength{\tabcolsep}{2pt}
\begin{table}[!htb]
	\begin{center}
		\caption{
			Results from CelebA-HQ test dataset, where Quick Draw dataset by Iskakov et al. \cite{iskakov2018semi} is used as masking method with mask hole-to-image ratios range between [0.01,0.6]. $\dagger$ Lower is better. $\uplus$ Higher is better.
		}
		\label{table:result1}
		\begin{tabular}{lrrrr}
			\hline
			Inpainting Method $\qquad$& FID $^\dagger$ & MAE $^\dagger$ & PSNR $^\uplus$ & SSIM $^\uplus$  \\
			\hline
			R-MNet-0.1  & 26.95 & 33.40 & 38.46 &0.88 \\
			Pathak et al. \cite{pathak2016context} & 29.96 & 123.54 & 32.61 & 0.69 \\
			Liu et al. \cite{liu2018image} & 15.86 & 98.01 & 33.03 & 0.81 \\
			Yu et al. \cite{yu2019free} & 4.29 & 43.10 & 39.69 & 0.92 \\  
			\textbf{R-MNet-0.4} & \textbf{3.09} & \textbf{31.91} & \textbf{40.40} &\textbf{0.94}\\
			\hline
		\end{tabular}
	\end{center}
\end{table}
\setlength{\tabcolsep}{2pt}
\begin{table}[!htb]
	\begin{center}
		\caption{
			The inpainting results of R-MNet-0.4 on Paris Street View and Places2, where Quick Draw dataset by Iskakov et al.\cite{iskakov2018semi} is used as masking method with mask hole-to-image ratios range between [0.01,0.6]. $\dagger$ Lower is better. $\uplus$ Higher is better.
		}
		\label{table:result3}
		\begin{tabular}{lrrrr}
			\hline
			Dataset $\qquad$& FID $^\dagger$ & MAE $^\dagger$ & PSNR $^\uplus$ & SSIM $^\uplus$  \\
			\hline
			Paris Street View & 17.64 & 33.81 & 39.55 & 0.91\\
			Places2  & 4.47 & 27.77 & 39.66 & 0.93\\
			\hline
		\end{tabular}
	\end{center}
\end{table}
A lower FID score indicates that the reconstructed images are close to the ground-truth. A similar judgement quantifies the MAE, though it measures the magnitude in pixel error between the ground-truth and reconstructed images. For PSNR and SSIM, higher values indicate good quality images closer to the ground-truth image. Looking at the results in Table~\ref{table:result1}, our model achieves better performances than other state-of-the-art methods. 

To test the effectiveness of our model, we conduct a test using the Nvidia Mask dataset \cite{liu2018image}, and show our results on the CelebA-HQ dataset in Figure~\ref{fig:results}. These masks are of different categories with hole-to-image area ratios. Note that these masks were not used during training of R-MNet. We used it for testing only to demonstrate our model superiority and robustness across mask. We carry out further experiments on Paris Street View and Places2 to generalize our model. Masks of the same sizes [0.01,0.6] are randomly selected during testing. Results can be seen on Table~\ref{table:result3} and Figure~\ref{fig:Paris-results}, which shows our model is able to generalize to various inpainting tasks and not just face inpainting. 

\subsection{Ablation Study}
\begin{figure}
	\centering
	\begin{subfigure}[b]{0.19\linewidth}        
		\centering
		\includegraphics[width=\linewidth]{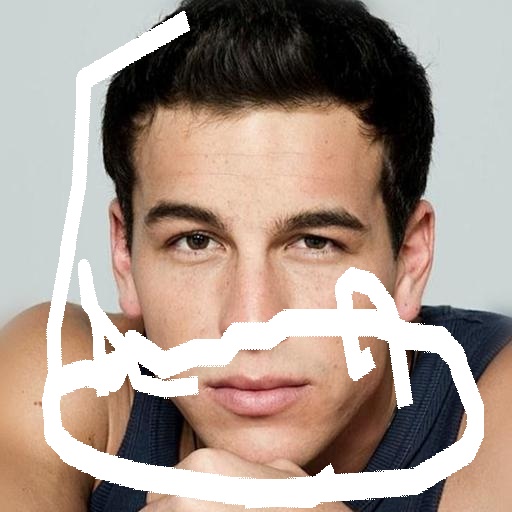}
	\end{subfigure}
	\begin{subfigure}[b]{0.19\linewidth}        
		\centering
		\includegraphics[width=\linewidth]{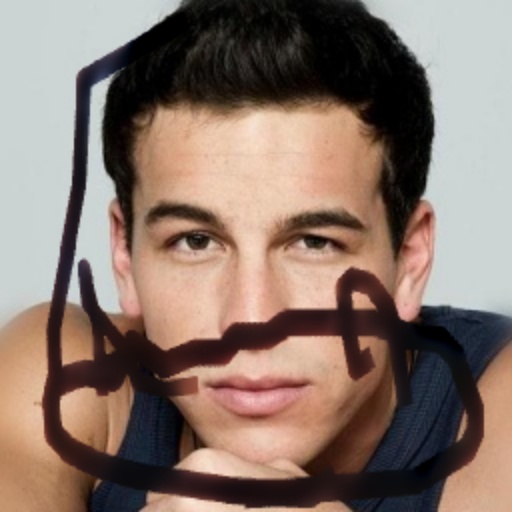}
	\end{subfigure}
	\begin{subfigure}[b]{0.19\linewidth}        
		\centering
		\includegraphics[width=\linewidth]{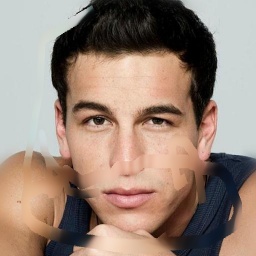}
	\end{subfigure}
	\begin{subfigure}[b]{0.19\linewidth}        
		\centering
		\includegraphics[width=\linewidth]{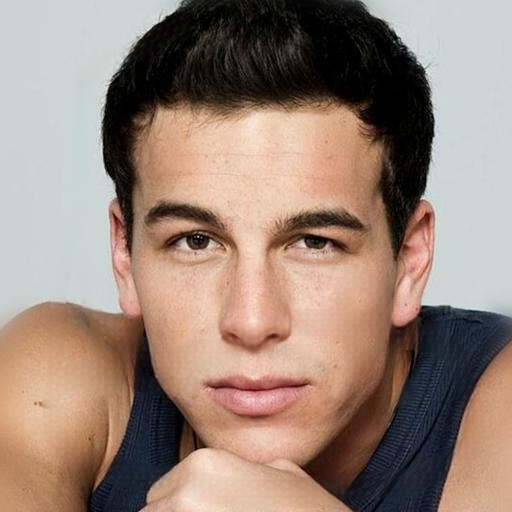}
	\end{subfigure}
	\begin{subfigure}[b]{0.19\linewidth}        
		\centering
		\includegraphics[width=\linewidth]{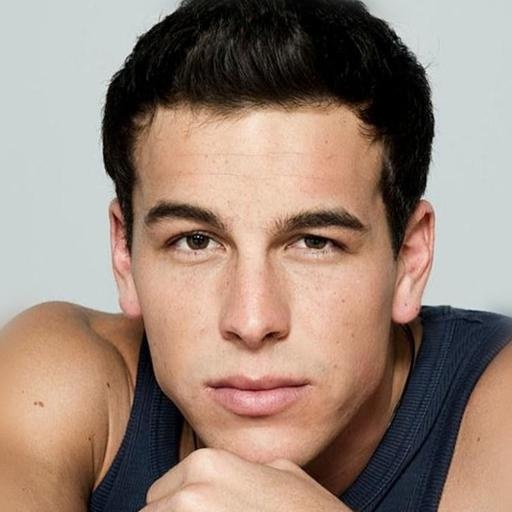}
	\end{subfigure}\\
	\begin{subfigure}[b]{0.19\linewidth}        
		\centering
		\includegraphics[width=\linewidth]{mask19.jpg}
		\caption{Masked}
	\end{subfigure}
	\begin{subfigure}[b]{0.19\linewidth}        
		\centering
		\includegraphics[width=\linewidth]{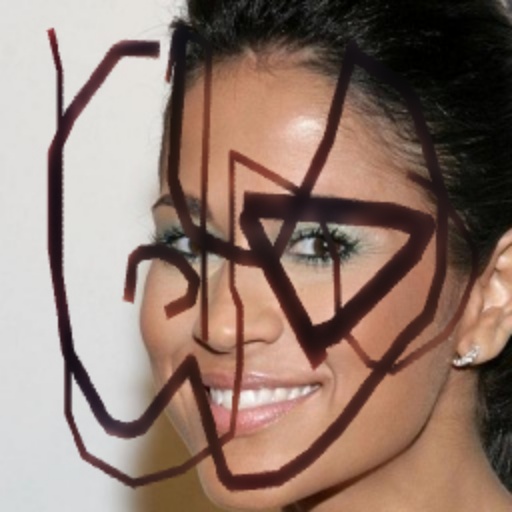}
		\caption{$\lambda = 0$}
	\end{subfigure}
	\begin{subfigure}[b]{0.19\linewidth}        
		\centering
		\includegraphics[width=\linewidth]{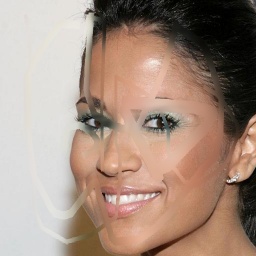}
		\caption{$\lambda = 0.1$}
	\end{subfigure}
	\begin{subfigure}[b]{0.19\linewidth}        
		\centering
		\includegraphics[width=\linewidth]{RMNET19.jpg}
		\caption{$\lambda = 0.4$}
	\end{subfigure}
	\begin{subfigure}[b]{0.19\linewidth}        
		\centering
		\includegraphics[width=\linewidth]{GT19.jpg}
		\caption{GT}
	\end{subfigure}\\	     
	\caption{Visual results on ablation study where (a) is the input masked image (b) results of the R-MNet-Base model without $\boldsymbol\ell_{rm}$. As loss on this model we used $\boldsymbol\ell_{2}$ and binary-cross-entropy as loss (c) R-MNet with $\boldsymbol\ell_{rm}$ loss at with weight application of 0.1  (d) R-MNet with full $\boldsymbol\ell_{VP}(\phi)$ with $\lambda$=0.4 on Quick-Draw \cite{iskakov2018semi} as masking method.}
	\label{fig:nresults}
\end{figure}

We investigate the effectiveness of reversed mask loss, and conduct experiments at different $\lambda$ and compare its performance using hole-to-image area ratios between [0.01, 0.6]. We use $\lambda$= \textbf{0, 0.1, 0.3, 0.4, 0.5} on reversed mask loss for different experiments with the same settings. The results are shown in Table~\ref{table:ablation}. When $\lambda$=\textbf{0} the model has no access to the reversed mask to compute the loss function, we realise that the mask residue is left on the image as shown on Figure~\ref{fig:nresults}.  Based on the output images, although the spatial information of the image is preserved, the convolutional inpainted regions need assistance to minimise the loss between the mask and the reverse mask loss during prediction. We start by giving a small value to $\lambda$=\textbf{0.1}, we notice that the results get better but we obtain poor performance visually. That is, the pixels in the mask region did not blend properly with the surrounding pixels, leaving the image with inconsistencies in texture. We experiment further with $\lambda$= \textbf{0.3, 0.4, 0.5} and carried out subjective evaluations using FID, MAE, PSNR and SSIM. With $\lambda$= \textbf{0.4}, we obtained the best results with no further improvement by increasing the value of $\lambda$. The mask as input to the CNN allows the network to learn the size of the corrupted region. The bigger the mask, the longer it takes to achieve perceptual similarity. This is because the region grows bigger and the network takes longer due to a smaller proportion of the loss covering the entire image to ensure the inpainted region is semantically consistent with the rest of the image.

\setlength{\tabcolsep}{2pt}
\begin{table}
	\begin{center}
		\caption{
			Results from Paris Street View and Places2 using Quick Draw dataset by Iskakov et al. \cite{iskakov2018semi} as masking method with mask hole-to-image ratios [0.01,0.6]. $\dagger$ Lower is better. $\uplus$ Higher is better. 
		}
		\label{table:ablation}
		\begin{tabular}{lrrrr}
			\hline\noalign{\smallskip}
			$\mathcal{L}_{rm}$ weight $\qquad\qquad$ & FID $^\dagger$ & MAE $^\dagger$ & PSNR $^\uplus$  & SSIM $^\uplus$  \\
			\noalign{\smallskip}
			\hline
			\noalign{\smallskip}
			$\lambda$=0.1, R-MNet-0.1 & 26.95 & 33.40 & 38.46 &0.88 \\
			$\lambda$=0.3, R-MNet-0.3 & 4.14 & 31.57 &40.20 &0.93 \\
			$\lambda$=0.4, \textbf{R-MNet-0.4} & \textbf{3.09} & \textbf{31.91} & \textbf{40.40} &\textbf{0.94}\\
			$\lambda$=0.5, R-MNet-0.5 & 4.14 & 31.0 & 40.0 &0.93 \\
			\hline
		\end{tabular}
	\end{center}
\end{table}

\section{Discussion}
\label{sec:discussion}
The ability to generalize good performance on machine learning algorithms based on end-to-end mapping of real data distribution to unseen data, is vital to the learning outcome by various models. In image inpainting, algorithms based on generative networks, predict  missing regions following a real data distribution from a large dataset. Typical approaches predict these hidden regions by applying an encoding-decoding process to the image, where the missing regions are defined usually by a binary mask. The encoding process will produce a high dimensional feature representation of the image where the missing information has been recovered, while the decoding process will generate the original image, i.e. the input image without missing information.

Generally the learning procedure of the model parameters is performed by solving a loss function minimization. Often, the model parameters are learned using a forward-backward process. In the forward pass the loss function calculates an error between latent distribution of real and generated data. The loss is then back-propagated into the model to update the parameters weight (backward pass).

The ability of our model to identify the missing regions of the input image is essentially assisted by our reverse mask loss, that will force the network to primarily focus on the prediction of the missing regions. The reverse mask loss combined with the perceptual loss and the preserved spatial information within the network, will ensure accurate prediction of missing regions while keeping the general structure of the image and high resolution details. 

A practice of existing inpainting GAN is the requirement to apply the mask on an image to obtain a composite image (masked-image), which in turn share the same pixel level. Additionally, when using irregular masks, these regions to be inpainted can be seen by the algorithm as a square bounding box that contains both visible and missing pixels, which can cause the GAN to generate images that can contain structural and textural artefacts.  

To overcome this issue, we modify the usual approach by having two inputs to our model, during the training, the image and its associated binary mask. This allow us to have access to the mask, at the end of our encoding/decoding network, through the spatial preserving operation and gives us the ability to compute the following, that will be used during the loss computation:
\begin{itemize}
	\item A reversed mask applied to the output image. 
	\item Use a spatial preserving operation to get the original masked-image.
	\item Use matrix operands to add the reversed mask image back on the masked-image.
\end{itemize}
By using this method of internal masking and restoring, our network can inpaint only the required features while maintaining the original image structure and texture with high level of details. Our network shows better achievement, when compared to state-of-the-art methods, numerically and visually, where the output image are visually closer to the original image than other approaches.

Globally we demonstrate through our approach that combining global (perceptual) and specific (reverse mask) loss we can achieve better performances, thus overcoming the limitation of having a model trained only using global loss.
\section{Conclusion}
\label{secc:conclusion}
In this paper, we propose a novel approach using Reverse Masking combined with Wasserstein GAN to perform image inpainting task on various binary mask shapes and sizes. Our model targets missing pixels and reconstructs an image with structural and textural consistency. We demonstrate that our model can perform accurate image inpainting of masked regions on high resolution images while preserving image details. Through our experimental results, we have shown that training our model alongside performing reverse matrix operands of the mask is beneficial to image inpainting. We also show that our model when compared with the state-of-the-art methods can obtain competitive results.

\section*{Acknowledgement}
This work was supported by The Royal Society UK (INF\textbackslash PHD\textbackslash 180007 and IF160006). We gratefully acknowledge the support of NVIDIA Corporation with the donation of the Quadro P6000 used for this research.

{\small
\bibliographystyle{ieee_fullname}
\bibliography{rmnetbib}

\begin{thebibliography}{10}\itemsep=-1pt

\bibitem{arjovsky2017wasserstein}
Martin Arjovsky, Soumith Chintala, and L{\'e}on Bottou.
\newblock Wasserstein generative adversarial networks.
\newblock In {\em International Conference on Machine Learning}.

\bibitem{barnes2009patchmatch}
Connelly Barnes, Eli Shechtman, Adam Finkelstein, and Dan~B Goldman.
\newblock Patchmatch: A randomized correspondence algorithm for structural
  image editing.
\newblock {\em ACM Transactions on Graphics}, 28(3):24, 2009.

\bibitem{bertalmio2000image}
Marcelo Bertalmio, Guillermo Sapiro, Vincent Caselles, and Coloma Ballester.
\newblock Image inpainting.
\newblock In {\em ACM Conference on Computer graphics and interactive
  techniques}, 2000.

\bibitem{criminisi2004region}
Antonio Criminisi, Patrick P{\'e}rez, and Kentaro Toyama.
\newblock Region filling and object removal by exemplar-based image inpainting.
\newblock {\em IEEE Transactions on image processing}, 13(9):1200--1212, 2004.

\bibitem{doersch2012makes}
Carl Doersch, Saurabh Singh, Abhinav Gupta, Josef Sivic, and Alexei Efros.
\newblock What makes paris look like paris?
\newblock {\em ACM Transactions on Graphics}, 31(4), 2012.

\bibitem{efros1999texture}
Alexei~A Efros and Thomas~K Leung.
\newblock Texture synthesis by non-parametric sampling.
\newblock In {\em International Conference on Computer Vision}, 1999.

\bibitem{gatys2016neural}
Leon Gatys, Alexander Ecker, and Matthias Bethge.
\newblock A neural algorithm of artistic style.
\newblock {\em Journal of Vision}, 16(12):326--326, 2016.

\bibitem{goodfellow2014generative}
Ian Goodfellow, Jean Pouget-Abadie, Mehdi Mirza, Bing Xu, David Warde-Farley,
  Sherjil Ozair, Aaron Courville, and Yoshua Bengio.
\newblock Generative adversarial nets.
\newblock In {\em Advances in neural information processing systems}, pages
  2672--2680, 2014.

\bibitem{guo2019progressive}
Zongyu Guo, Zhibo Chen, Tao Yu, Jiale Chen, and Sen Liu.
\newblock Progressive image inpainting with full-resolution residual network.
\newblock In {\em ACM International Conference on Multimedia}, 2019.

\bibitem{he2016deep}
Kaiming He, Xiangyu Zhang, Shaoqing Ren, and Jian Sun.
\newblock Deep residual learning for image recognition.
\newblock In {\em IEEE conference on computer vision and pattern recognition},
  2016.

\bibitem{heusel2017gans}
Martin Heusel, Hubert Ramsauer, Thomas Unterthiner, Bernhard Nessler, and Sepp
  Hochreiter.
\newblock Gans trained by a two time-scale update rule converge to a local nash
  equilibrium.
\newblock In {\em Advances in neural information processing systems}, pages
  6626--6637, 2017.

\bibitem{huang2019image}
Ying Huang, Maorui Wang, Ying Qian, Shuohao Lin, and Xiaohan Yang.
\newblock Image completion based on gans with a new loss function.
\newblock {\em Journal of Physics: Conference Series}, 1229(1):012030, 2019.

\bibitem{iizuka2017globally}
Satoshi Iizuka, Edgar Simo-Serra, and Hiroshi Ishikawa.
\newblock Globally and locally consistent image completion.
\newblock {\em ACM Transactions on Graphics}, 36(4):107, 2017.

\bibitem{iskakov2018semi}
Karim Iskakov.
\newblock Semi-parametric image inpainting.
\newblock {\em arXiv preprint arXiv:1807.02855}, 2018.

\bibitem{jain2007supervised}
Viren Jain, Joseph~F Murray, Fabian Roth, Srinivas Turaga, Valentin Zhigulin,
  Kevin~L Briggman, Moritz~N Helmstaedter, Winfried Denk, and H~Sebastian
  Seung.
\newblock Supervised learning of image restoration with convolutional networks.
\newblock In {\em IEEE International Conference on Computer Vision}, 2007.

\bibitem{jam2020symmetric}
Jireh Jam, Connah Kendrick, Vincent Drouard, Kevin Walker, Gee-Sern Hsu, and
  Moi~Hoon Yap.
\newblock Symmetric skip connection wasserstein gan for high-resolution facial
  image inpainting.
\newblock {\em arXiv preprint arXiv:2001.03725}, 2020.

\bibitem{johnson2016perceptual}
Justin Johnson, Alexandre Alahi, and Li Fei-Fei.
\newblock Perceptual losses for real-time style transfer and super-resolution.
\newblock In {\em European conference on computer vision}, 2016.

\bibitem{karras2018progressive}
Tero Karras, Timo Aila, Samuli Laine, and Jaakko Lehtinen.
\newblock Progressive growing of gans for improved quality, stability, and
  variation.
\newblock In {\em International Conference on Learning Representations}, 2018.

\bibitem{kingma2015adam}
Diederik~P Kingma and Jimmy~Lei Ba.
\newblock Adam: A method for stochastic gradient descent.
\newblock In {\em International Conference on Learning Representations}, 2015.

\bibitem{krizhevsky2012imagenet}
Alex Krizhevsky, Ilya Sutskever, and Geoffrey~E Hinton.
\newblock Imagenet classification with deep convolutional neural networks.
\newblock In {\em Advances in neural information processing systems}, pages
  1097--1105, 2012.

\bibitem{li2019progressive}
Jingyuan Li, Fengxiang He, Lefei Zhang, Bo Du, and Dacheng Tao.
\newblock Progressive reconstruction of visual structure for image inpainting.
\newblock In {\em IEEE International Conference on Computer Vision}, 2019.

\bibitem{li2017generative}
Yijun Li, Sifei Liu, Jimei Yang, and Ming-Hsuan Yang.
\newblock Generative face completion.
\newblock In {\em IEEE Conference on Computer Vision and Pattern Recognition},
  2017.

\bibitem{liu2018image}
Guilin Liu, Fitsum~A Reda, Kevin~J Shih, Ting-Chun Wang, Andrew Tao, and Bryan
  Catanzaro.
\newblock Image inpainting for irregular holes using partial convolutions.
\newblock In {\em European Conference on Computer Vision}, 2018.

\bibitem{liu2019coherent}
Hongyu Liu, Bin Jiang, Yi Xiao, and Chao Yang.
\newblock Coherent semantic attention for image inpainting.
\newblock In {\em IEEE International Conference on Computer Vision}, 2019.

\bibitem{liu2015deep}
Ziwei Liu, Ping Luo, Xiaogang Wang, and Xiaoou Tang.
\newblock Deep learning face attributes in the wild.
\newblock In {\em IEEE International Conference on Computer Vision}, 2015.

\bibitem{pathak2016context}
Deepak Pathak, Philipp Krahenbuhl, Jeff Donahue, Trevor Darrell, and Alexei~A
  Efros.
\newblock Context encoders: Feature learning by inpainting.
\newblock In {\em IEEE Conference on Computer Vision and Pattern Recognition},
  2016.

\bibitem{ren2019structureflow}
Yurui Ren, Xiaoming Yu, Ruonan Zhang, Thomas~H Li, Shan Liu, and Ge Li.
\newblock Structureflow: Image inpainting via structure-aware appearance flow.
\newblock In {\em IEEE International Conference on Computer Vision}, 2019.

\bibitem{ronneberger2015u}
Olaf Ronneberger, Philipp Fischer, and Thomas Brox.
\newblock U-net: Convolutional networks for biomedical image segmentation.
\newblock In {\em International Conference on Medical image computing and
  computer-assisted intervention}, 2015.

\bibitem{rosebrock2019deep}
Adrian Rosebrock.
\newblock {\em Deep Learning for Computer Vision with Python}.
\newblock PyImageSearch.com, 2.1.0 edition, 2019.

\bibitem{shi2016real}
Wenzhe Shi, Jose Caballero, Ferenc Husz{\'a}r, Johannes Totz, Andrew~P Aitken,
  Rob Bishop, Daniel Rueckert, and Zehan Wang.
\newblock Real-time single image and video super-resolution using an efficient
  sub-pixel convolutional neural network.
\newblock In {\em IEEE conference on computer vision and pattern recognition},
  2016.

\bibitem{simonyan2015very}
Karen Simonyan and Andrew Zisserman.
\newblock Very deep convolutional networks for large-scale image recognition.
\newblock In {\em International Conference on Learning Representations}, 2015.

\bibitem{wang2019musical}
Ning Wang, Jingyuan Li, Lefei Zhang, and Bo Du.
\newblock Musical: Multi-scale image contextual attention learning for
  inpainting.
\newblock In {\em IJCAI}, pages 3748--3754, 2019.

\bibitem{wang2020multistage}
Ning Wang, Sihan Ma, Jingyuan Li, Yipeng Zhang, and Lefei Zhang.
\newblock Multistage attention network for image inpainting.
\newblock {\em Pattern Recognition}, page 107448, 2020.

\bibitem{wang2019laplacian}
Qiang Wang, Huijie Fan, Gan Sun, Yang Cong, and Yandong Tang.
\newblock Laplacian pyramid adversarial network for face completion.
\newblock {\em Pattern Recognition}, 88:493--505, 2019.

\bibitem{wang2004image}
Zhou Wang, Alan~C Bovik, Hamid~R Sheikh, Eero~P Simoncelli, et~al.
\newblock Image quality assessment: from error visibility to structural
  similarity.
\newblock {\em IEEE transactions on image processing}, 13(4):600--612, 2004.

\bibitem{xie2019image}
Chaohao Xie, Shaohui Liu, Chao Li, Ming-Ming Cheng, Wangmeng Zuo, Xiao Liu,
  Shilei Wen, and Errui Ding.
\newblock Image inpainting with learnable bidirectional attention maps.
\newblock In {\em Proceedings of the IEEE International Conference on Computer
  Vision}, pages 8858--8867, 2019.

\bibitem{yan2018shift}
Zhaoyi Yan, Xiaoming Li, Mu Li, Wangmeng Zuo, and Shiguang Shan.
\newblock Shift-net: Image inpainting via deep feature rearrangement.
\newblock In {\em European Conference on Computer Vision}, 2018.

\bibitem{yang2017high}
Chao Yang, Xin Lu, Zhe Lin, Eli Shechtman, Oliver Wang, and Hao Li.
\newblock High-resolution image inpainting using multi-scale neural patch
  synthesis.
\newblock In {\em IEEE Conference on Computer Vision and Pattern Recognition},
  2017.

\bibitem{yeh2017semantic}
Raymond~A Yeh, Chen Chen, Teck-Yian Lim, Alexander~G Schwing, Mark
  Hasegawa-Johnson, and Minh~N Do.
\newblock Semantic image inpainting with deep generative models.
\newblock In {\em IEEE Conference on Computer Vision and Pattern Recognition},
  2017.

\bibitem{yu2018generative}
Jiahui Yu, Zhe Lin, Jimei Yang, Xiaohui Shen, Xin Lu, and Thomas~S Huang.
\newblock Generative image inpainting with contextual attention.
\newblock In {\em IEEE conference on computer vision and pattern recognition},
  2018.

\bibitem{yu2019free}
Jiahui Yu, Zhe Lin, Jimei Yang, Xiaohui Shen, Xin Lu, and Thomas~S Huang.
\newblock Free-form image inpainting with gated convolution.
\newblock In {\em IEEE International Conference on Computer Vision}, 2019.

\bibitem{zeng2019learning}
Yanhong Zeng, Jianlong Fu, Hongyang Chao, and Baining Guo.
\newblock Learning pyramid-context encoder network for high-quality image
  inpainting.
\newblock In {\em IEEE Conference on Computer Vision and Pattern Recognition},
  2019.

\bibitem{zhou2018places}
Bolei Zhou, Agata Lapedriza, Aditya Khosla, Aude Oliva, and Antonio Torralba.
\newblock Places: A 10 million image database for scene recognition.
\newblock {\em IEEE transactions on pattern analysis and machine intelligence},
  40(6):1452--1464, 2018.

\end{thebibliography}
}

\end{document}